\documentclass{article}

\usepackage[final]{neurips_2024}

\usepackage[utf8]{inputenc} 
\usepackage[T1]{fontenc}    
\usepackage{hyperref}       
\usepackage{url}            
\usepackage{booktabs}       
\usepackage{amsfonts}       
\usepackage{nicefrac}       
\usepackage{microtype}      
\usepackage{xcolor}         
\usepackage{amsmath}
\usepackage{comment}
\usepackage{graphicx} 
\usepackage{caption}
\usepackage{subcaption}

\title{Do Language Models Have Bayesian Brains? Distinguishing Stochastic and Deterministic Decision Patterns within Large Language Models}

\author{%
  Andrea Yaoyun Cui\thanks{Equal Contribution} \\
  University of Illinois Urbana Champaign\\
  \texttt{yaoyunc2@illinois.edu} \\
  \And
  Pengfei Yu\textsuperscript{\footnotemark[1]}\\
  Boson AI \\
  \texttt{pengfei@boson.ai} \\
}

\begin{document}

\maketitle

\begin{abstract}
Language models are essentially probability distributions over token sequences. Auto-regressive models generate sentences by iteratively computing and sampling from the distribution of the next token. This iterative sampling introduces stochasticity, leading to the assumption that language models make probabilistic decisions, similar to sampling from unknown distributions. Building on this assumption, prior research has used simulated Gibbs sampling, inspired by experiments designed to elicit human priors, to infer the priors of language models. In this paper, we revisit a critical question: Do language models possess Bayesian brains? Our findings show that under certain conditions, language models can exhibit near-deterministic decision-making, such as producing maximum likelihood estimations, even with a non-zero sampling temperature. This challenges the sampling assumption and undermines previous methods for eliciting human-like priors. Furthermore, we demonstrate that without proper scrutiny, a system with deterministic behavior undergoing simulated Gibbs sampling can converge to a "false prior." To address this, we propose a straightforward approach to distinguish between stochastic and deterministic decision patterns in Gibbs sampling, helping to prevent the inference of misleading language model priors. We experiment on a variety of large language models to identify their decision patterns under various circumstances. Our results provide key insights in understanding decision making of large language models.
\end{abstract}
\section{Introduction}

Auto-regressive models, the backbone architecture of prevailing large language models~(LLMs), generate text by iteratively sampling the next token in a sequence. This has led to the common assumption that language models make stochastic decisions, meaning they sample from non-trivial distributions, as opposed to deterministic decisions, where choices are fully determined by the current states or the inputs. Building upon this assumption, methods such as simulated Gibbs sampling have been used to infer language model priors~\cite{zhu2024eliciting,zhu2024recovering}, drawing parallels to experiments that elicit human priors~\cite{griffiths2007language,kalish2007iterated,sanborn2007markov,griffiths2008using,canini2014revealing,yeung2015identifying,reali2009evolution,lewandowsky2009wisdom}. Understanding priors of LLMs can provide useful insights in aligning them with human preference, which is a key challenge in ensuring safety and helpfulness.

However, we question whether language models always make stochastic decisions. Our findings suggest that under certain conditions, models exhibit near-deterministic behaviors, e.g., converging to maximum likelihood estimates, even with non-zero sampling temperatures. This challenges the belief that language models sample from distributions, raising concerns about the validity of inferred priors.

To better understand the nature of decision-making in LLMs, we propose a method to distinguish stochastic from deterministic decision-making in language models by analyzing their responses under varied initial conditions. Our experiments show that many models switch between stochastic and deterministic patterns, offering deeper insights into their behavior. These findings are crucial for accurately interpreting language models' behavior and priors, particularly in contexts where aligning model behavior with human preferences is crucial for safety and effectiveness.

\section{Related Work}
Iterated learning has been widely used in psychology to elicit human priors~\cite{griffiths2007language,kalish2007iterated,sanborn2007markov,griffiths2008using,canini2014revealing,yeung2015identifying,reali2009evolution,lewandowsky2009wisdom,yamakoshi-etal-2022-probing}. In this approach, a specific variable \(\theta\) is linked to an observational variable \(\omega\), and participants estimate \(\theta_i\) based on an observation \(\omega_i\) drawn from the distribution \(p(\cdot|\theta_{i-1})\), where \(\theta_{i-1}\) is the estimate from the previous round. Assuming participants use a shared posterior distribution \(p(\theta|\omega) \propto p(\omega|\theta)p(\theta)\), this process mirrors Gibbs sampling~\cite{geman1984stochastic}, eventually driving \(\theta_i\) to converge to the distribution \(p(\theta)\). For a sufficiently large \(N\), the sequence \(\{\theta_n\}_{n>N}\) approximates the human prior on \(\theta\). 

This iterative framework has also been applied to eliciting priors from large language models on causal strength, proportion estimation, everyday quantities~\cite{zhu2024eliciting} and mental representations~\cite{zhu2024recovering}. 
\section{Analysis}\label{sec:app}
In this section, we present a case study on proportion estimation in large language models based on \cite{zhu2024eliciting}. By examining the application of iterated learning in this scenario, we highlight the similarities and differences between stochastic and deterministic decision-making patterns. Building on this, we present a methodology for identifying decision patterns in large language models.

\subsection{Maximum Likelihood Proportion Estimation}\label{sec:mle}
Proportion estimation typically deals with binary outcomes, such as estimating the probability \(\theta\) of a coin landing heads~\cite{reali2009evolution}. In \cite{zhu2024eliciting}, this method is adapted to large language models to estimate \(p(\theta)\) as follows:
\begin{enumerate}
    \item Begin with a random sample of heads \(\omega_0\) from \(N\) coin tosses.
    \item Iteratively prompt the language model to estimate the number of heads \(\Omega_i\) from \(M\) coin tosses, given the prior result \(\omega_{i-1}\) heads from \(N\) tosses.
    \item Compute \(\theta_i = \Omega_i / M\), then sample \(\omega_i\sim \text{Binomial}(N, \theta_i)\).
\end{enumerate}

The language model produces a bimodal prior distribution over \(\theta\), concentrating at the extremes \(\theta = 0\) and \(\theta = 1\), as reported in \cite{zhu2024eliciting}. This outcome is consistent with findings on human priors~\cite{reali2009evolution} but is counterintuitive: why would large language models assume that coins can only be totally biased? In this work, we propose an alternative explanation for the bimodal result by assuming a deterministic decision process based on maximum likelihood estimation (MLE) of \(\theta\), where \(\Omega_i = M \frac{\omega_{i-1}}{N}\). Under this assumption, the language models no longer sample from posterior distributions, and the iterated learning process no longer functions as Gibbs sampling. To analyze the convergence of such a process, we model the stochastic process \(\{\omega_i\}\), where the randomness arises from sampling \(\omega_i\sim \text{Binomial}(N, \theta_i)\). It is evident that \(\{\omega_i\}\) forms a Markov process with the following transition:
\begin{equation}
    \omega_i \sim \text{Binomial}(N, \frac{\omega_{i-1}}{N}).
\end{equation}

This transition matrix has notable features. First, \(\omega_i = 0\) and \(\omega_i = N\) are the only absorbing states, meaning the stationary distribution \(\hat{p}_{\text{MLE}}\) has a support set of only \(\{0, N\}\). Second, we have:
\begin{equation}
    \mathbb{E}[\omega_i] = \mathbb{E}_{\omega_{i-1}}\left[N \times \frac{\omega_{i-1}}{N}\right] = \mathbb{E}[\omega_{i-1}],
\end{equation}
which leads to:
\begin{equation}
    \mathbb{E}[\omega_0] = \mathbb{E}_{\omega \sim \hat{p}_{\text{MLE}}}[\omega] = \hat{p}(N) N.
\end{equation}

Thus, the stationary distribution for an MLE decision maker is given by:
\begin{equation}
    \hat{p}_{\text{MLE}}(0) = 1 - \frac{\mathbb{E}[\omega_0]}{N},\quad \hat{p}_{\text{MLE}}(N) = \frac{\mathbb{E}[\omega_0]}{N}.\label{eq:MLE}
\end{equation}
Since \(\theta_i = \frac{\omega_{i-1}}{N}\), the resulting distribution over \(\theta\) is also bimodal. If \(\frac{\mathbb{E}[\omega_0]}{N} = 0.5\), \(\hat{p}_{\text{MLE}}\) is symmetric around \(\theta = 0.5\), resembling a Beta distribution \(\text{B}(\alpha, \beta)\) with small \(\alpha\) and \(\beta\), as seen in~\cite{reali2009evolution, zhu2024eliciting}.

The key difference between the MLE decision process and the posterior sampling process lies in the absence of true prior elicitation in the MLE framework. Since \(\theta_i\) is not sampled from a posterior distribution \(p(\theta|\omega) \propto p(\omega|\theta)p(\theta)\), the stationary distribution reflects polarization induced by the iterative procedure rather than the model’s prior beliefs. This distinction underscores the need for caution when interpreting iterated learning results from LLMs, as they may not represent true priors.

\subsection{Distinguishing Stochastic and Deterministic Decision Making}
Given the existence of deterministic decision-making processes that yield similar results to posterior sampling, it is important to identify methods to differentiate between them in iterated learning. From the analysis above, we observe that in deterministic processes like MLE, the stationary distribution depends on the initial value \(\mathbb{E}[\omega_0]\). In contrast, in stochastic processes based on Gibbs sampling~\cite{geman1984stochastic}, the stationary distribution remains invariant across reasonable variations in \(\omega_0\), always converging to the same prior distribution over \(\theta\). This provides a simple criterion for distinguishing deterministic from stochastic decision-making mechanisms:
\begin{quote}
    \textit{Vary the initial value \(\omega_0\) and compare the resulting stationary distributions. If the distribution remains consistent across different \(\omega_0\) values, the process is likely stochastic. If not, it is more likely deterministic.}
\end{quote}
Although one might argue that some deterministic processes could converge to the same distribution regardless of \(\omega_0\), empirical evidence suggests that this typically signals the presence of specific priors, as we will demonstrate in Section~\ref{sec:exp}.

\section{Experiments}\label{sec:exp}
\subsection{Decision Making Settings and Models}

We consider two settings in~\cite{zhu2024eliciting}, following their prompting procedures:

\noindent\textbf{CoinFlip}  This is the process we analyze in Section~\ref{sec:app}. We use $N=10$ and $M=100$. We experiment with $\Omega_0\in\left[0,10\right]$ to distinguish between stochastic and deterministic decisions.

\noindent\textbf{LifeExpectancy} We aim to elicit priors of language models on life expectancy. To achieve this, we iteratively prompt language models to estimate the life expectancy $L_{i}$ of a random person, given the current age $A_{i-1}$ of this person. The next ``current age'' is sampled uniformly from $[1, L_i]$. 

We experiment with two open-source models: \texttt{LLaMA-3.1-70B-Instruct}~\cite{dubey2024llama} and \texttt{Gemma-2-2B-it}~\cite{team2024gemma}. For closed-source models, we use \texttt{gpt-4o-mini}~\cite{gpt4omini}, \texttt{gpt-4o}~\cite{achiam2023gpt,gpt4o}, \texttt{claude-3-haiku}~\cite{claude3} and \texttt{claude-3-5-sonnet}~\cite{claude3-5-sonnet}\footnote{We provide version numbers in Appendix}. 
We conduct all experiments with temperatures set to $1.0$.

\subsection{Results}
\begin{figure}[ht]
    \centering
    \begin{minipage}{0.32\textwidth}
        \centering
        \includegraphics[width=\linewidth]{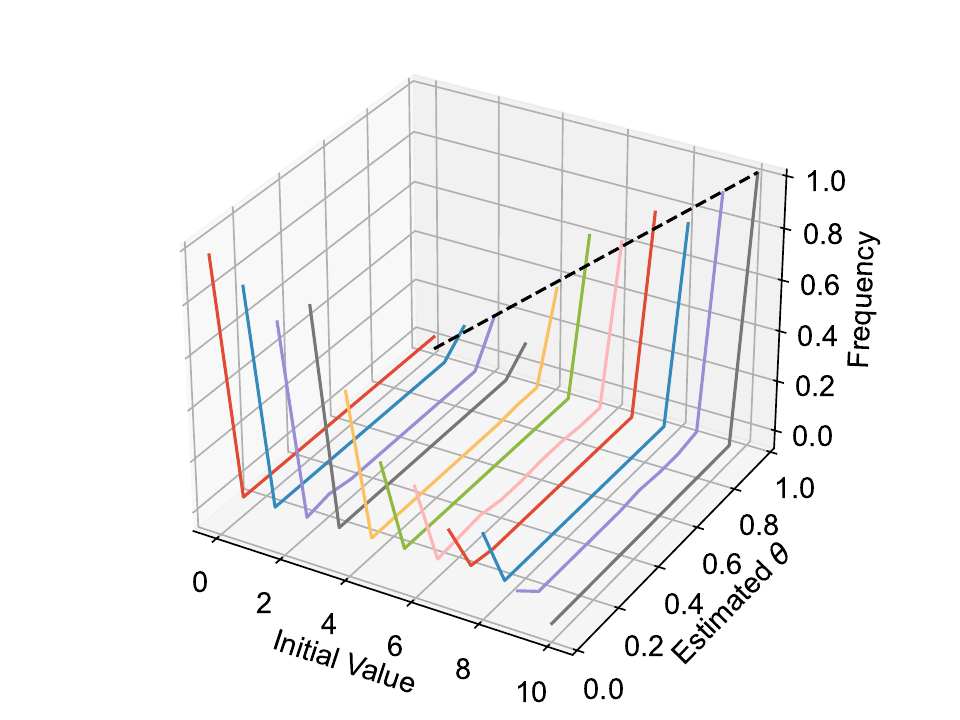}
        \subcaption{\texttt{gpt-4o-mini}}\label{fig:gpt4o}
    \end{minipage}
    \hfill
    \begin{minipage}{0.32\textwidth}
        \centering
        \includegraphics[width=\linewidth]{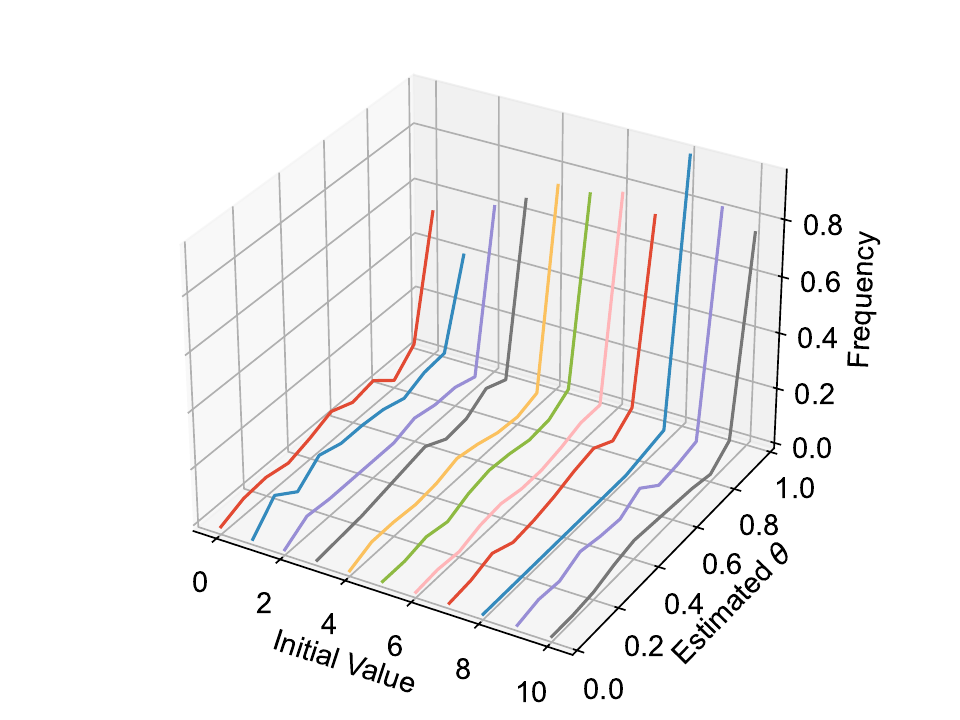}
        \subcaption{\texttt{gpt-4o}}
    \end{minipage}
    \hfill
    \begin{minipage}{0.32\textwidth}
        \centering
        \includegraphics[width=\linewidth]{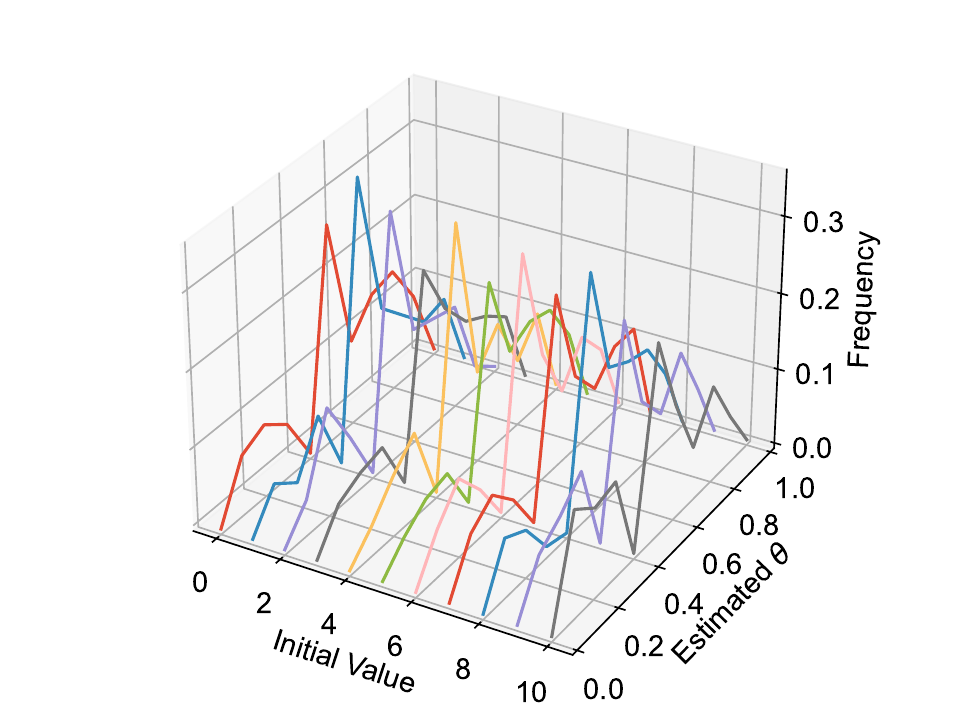}
        \subcaption{\texttt{claude-3-haiku}}
    \end{minipage}
    
    \begin{minipage}{0.32\textwidth}
        \centering
        \includegraphics[width=\linewidth]{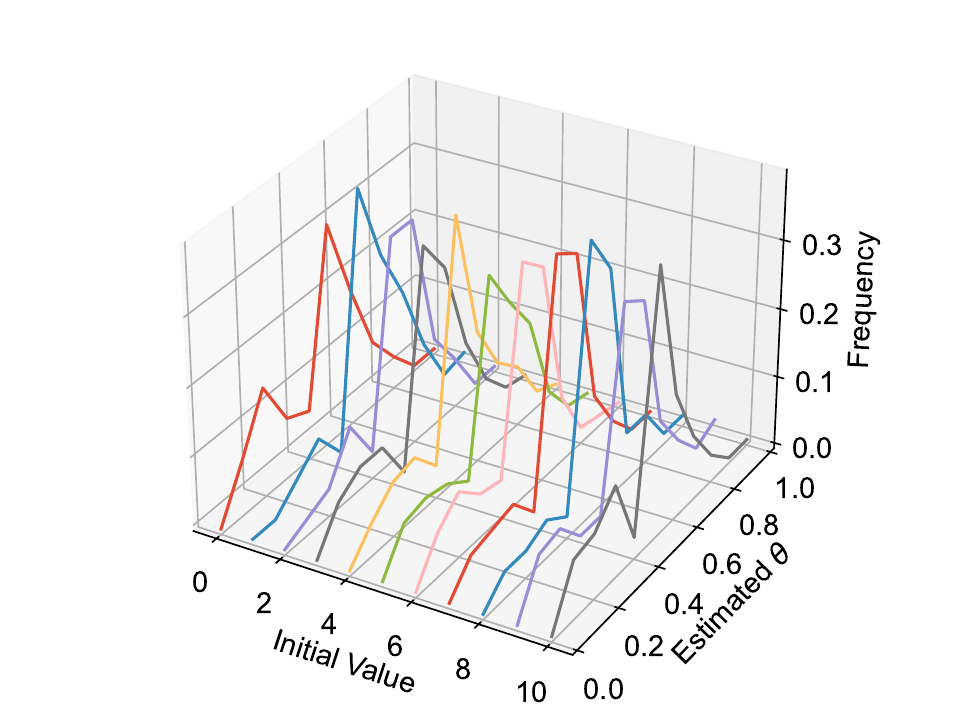}
        \subcaption{\texttt{claude-3-5-sonnet}}
    \end{minipage}
    \hfill
    \begin{minipage}{0.32\textwidth}
        \centering
        \includegraphics[width=\linewidth]{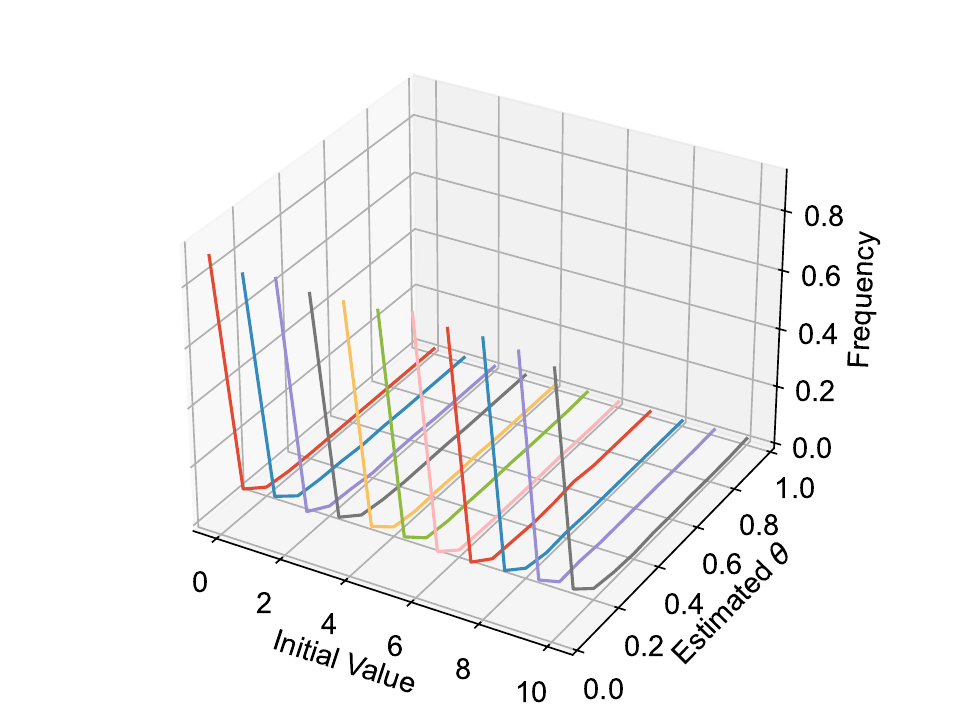}
        \subcaption{\texttt{Gemma-2-2b-i}t}
    \end{minipage}
    \hfill
    \begin{minipage}{0.32\textwidth}
        \centering
        \includegraphics[width=\linewidth]{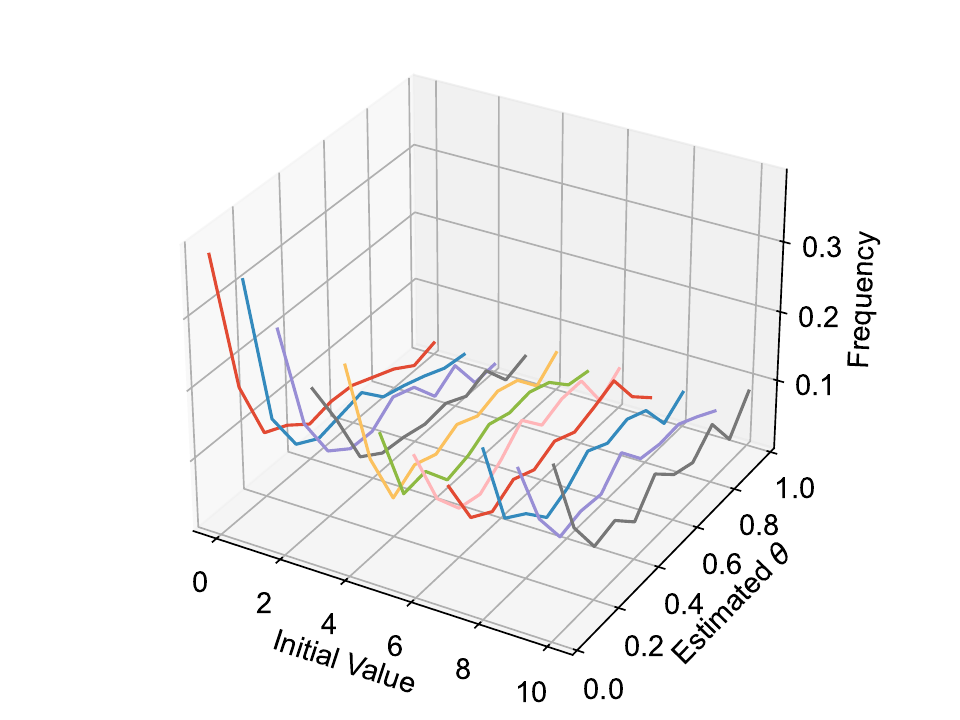}
        \subcaption{\texttt{Llama-3.1-70B-Instruct}}
    \end{minipage}
    \caption{\textbf{CoinFlip} results on all models.}\label{fig:coin}
\end{figure}
\begin{figure}[ht]
    \centering
    \begin{minipage}{0.32\textwidth}
        \centering
        \includegraphics[width=\linewidth]{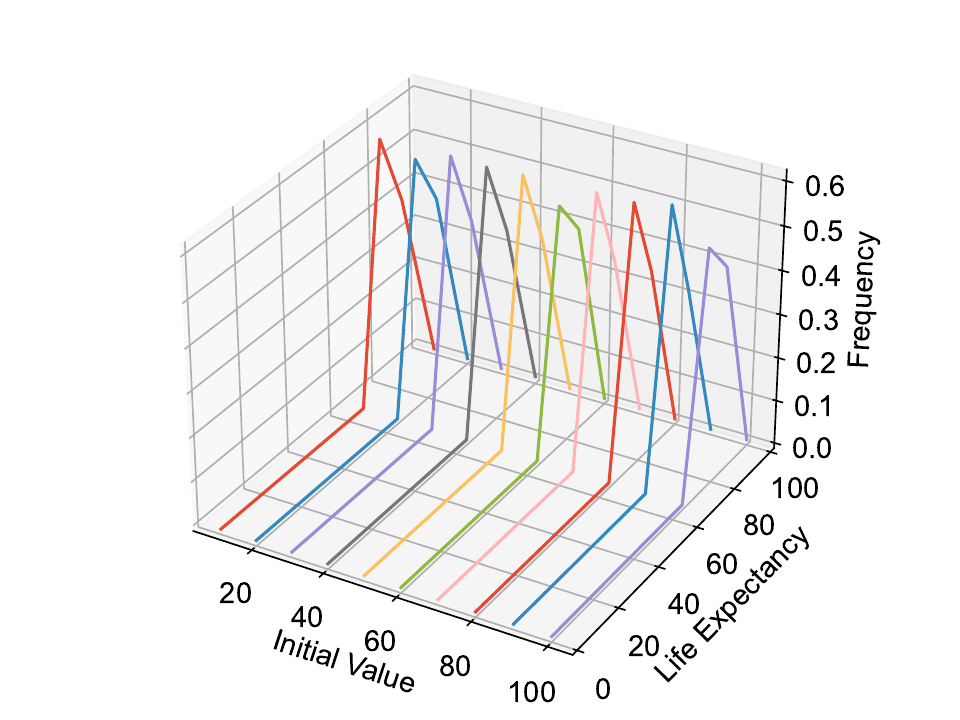}
        \subcaption{\texttt{gpt-4o-mini}}
    \end{minipage}
    \hfill
    \begin{minipage}{0.32\textwidth}
        \centering
        \includegraphics[width=\linewidth]{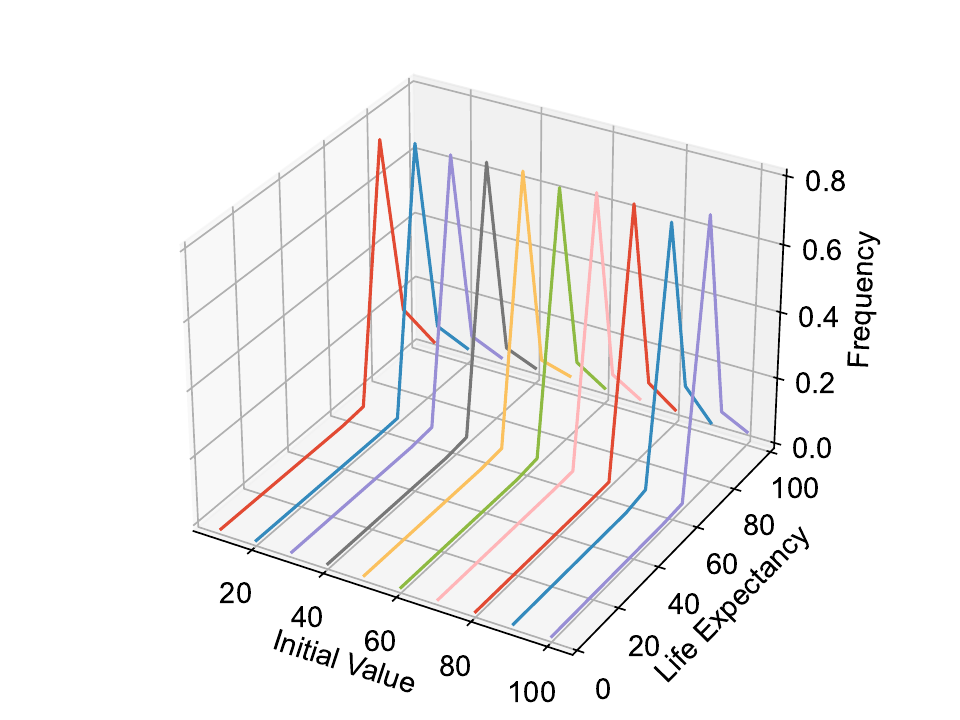}
        \subcaption{\texttt{Gemma-2-2b-i}t}
    \end{minipage}
    \hfill
    \begin{minipage}{0.32\textwidth}
        \centering
        \includegraphics[width=\linewidth]{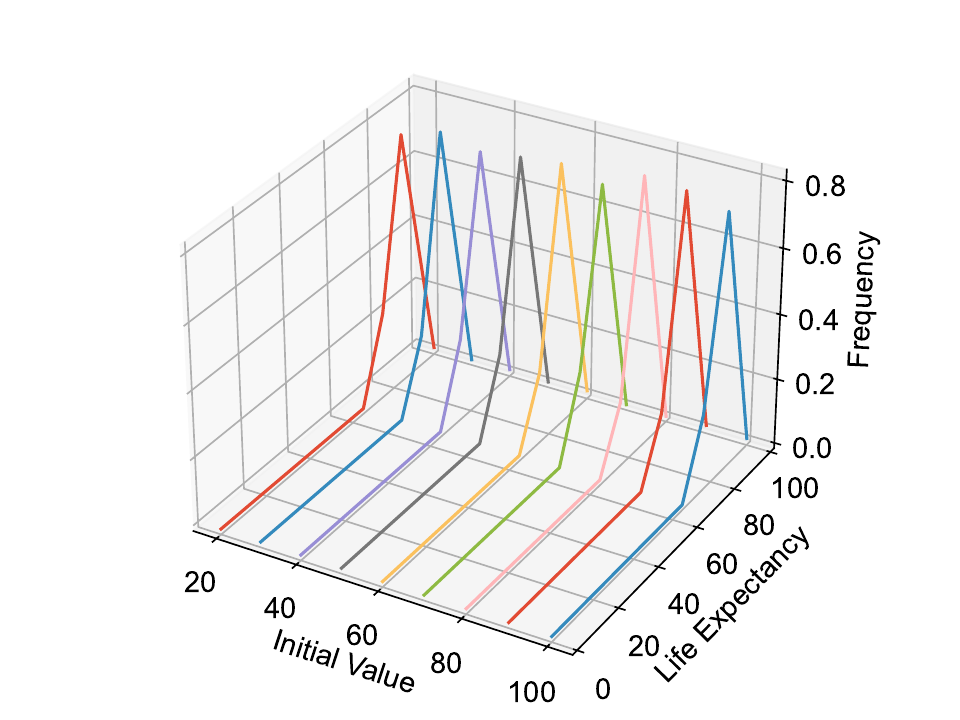}
        \subcaption{\texttt{claude-3-haiku}}
    \end{minipage}
    
    \caption{\textbf{LifeExpectancy} results on selected models. Results of other models in Appendix.}\label{fig:life}
\end{figure}

Figure~\ref{fig:coin} and Figure~\ref{fig:life} present the main results on a selection of models. We leave additional results in the Appendix due to space limits. Our key observations are:

\noindent\textbf{Variations in Decision Patterns on CoinFlip} 
The models show significant differences in their decision patterns for the CoinFlip task. \texttt{gpt-4o-mini} follows a deterministic decision-making pattern, with its distributions diverging across different initial values. The black dotted line lying in the plane of $\theta=1$ in Figure~\ref{fig:gpt4o}, which we fit featuring $p(\theta=1|\Omega_0)$ with respect to $\Omega_0$, also coincides with the MLE estimation in Equation~(\ref{eq:MLE}). On the other hand, \texttt{claude-3-5-sonnet}, \texttt{LLaMA-3.1-70B-Instruct}, and \texttt{claude-3-haiku} display stochastic behaviors, producing consistent distributions regardless of the initial values. These distributions, however, do not follow bimodal patterns concentrated at the extremes. Specifically, \texttt{claude-3-5-sonnet} and \texttt{claude-3-haiku} show normal-like distributions centered around \(\theta = 0.5\), while \texttt{LLaMA-3.1-70B-Instruct} skews towards the lower end with an exponential-like distribution.

Notably, \texttt{gpt-4o} consistently converges to a distribution focused at \(\theta = 1\). Further analysis reveals that \texttt{gpt-4o} actively avoids predicting 0 heads, which prevents \(\Omega = 0\) from becoming an absorbing state in the Markov process discussed in Section~\ref{sec:mle}. This behavior suggests that the model incorporates a prior that avoids \(\theta = 0\). However, it’s important to note that \texttt{gpt-4o}'s output of nearly \(p(\theta = 1) \approx 1\) reflects the iterative prompting process rather than a prior belief about biased coins. We observe similar behavior in \texttt{Gemma-2-2b-it}, though this model tends to avoid \(\theta = 1\) instead. An alternative explanation of these behaviors is, \texttt{gpt-4o} and \texttt{Gemma-2-2b-it} are both near-deterministic, except under certain conditions (e.g., when observed coins are all heads or tails).

\paragraph{Stochastic Decision Making on LifeExpectancy} 
In the \textbf{LifeExpectancy} task, all models consistently converge to similar distributions, demonstrating stochastic decision-making. The MLE estimate for \(L_i\) in this scenario would be \(A_{i-1}\), the current age, since the likelihood of encountering someone at age \(A_{i-1}\) is \(\frac{1}{L_i}\) for \(L_i \geq A_{i-1}\). Despite this, the models demonstrate enough ``intelligence'' to avoid naïve MLE estimations where life expectancy equals the observed age. 
\section{Conclusion and Discussion}
In this work, we revisited the question of whether language models consistently make stochastic decisions. Our theoretical analysis shows that deterministic decision processes can produce results similar to stochastic processes in iterated learning, though they differ in key underlying mechanisms. We proposed a method to distinguish between these decision patterns and conducted experiments across various models and scenarios to enhance our understanding of language models' decision-making behaviors.
\section{Limitation}
Our exploration is limited to two representative tasks in \citep{zhu2024eliciting} due to the cost of experimenting with large language models. We leave further evaluations on the generalizability of our approach to other tasks for future work. However, because Gibbs sampling guarantees convergence when the initial value falls in a reasonable range, our proposed identification approach is a sufficient condition to detect non-Gibbs-sampling processes, which further indicates non-stochastic decision processes in large language models. Therefore, our proposed approach demonstrates universal applicability to other tasks.

\bibliographystyle{plain}
\bibliography{citations}

\begin{thebibliography}{10}

\bibitem{achiam2023gpt}
Josh Achiam, Steven Adler, Sandhini Agarwal, Lama Ahmad, Ilge Akkaya, Florencia~Leoni Aleman, Diogo Almeida, Janko Altenschmidt, Sam Altman, Shyamal Anadkat, et~al.
\newblock Gpt-4 technical report.
\newblock {\em arXiv preprint arXiv:2303.08774}, 2023.

\bibitem{claude3}
Anthropic.
\newblock The claude 3 model family: Opus, sonnet, haiku, 2024.
\newblock Accessed: 13-Sep-2024.

\bibitem{claude3-5-sonnet}
Anthropic.
\newblock Claude 3.5 sonnet, 2024.
\newblock Accessed: 13-Sep-2024.

\bibitem{canini2014revealing}
Kevin~R Canini, Thomas~L Griffiths, Wolf Vanpaemel, and Michael~L Kalish.
\newblock Revealing human inductive biases for category learning by simulating cultural transmission.
\newblock {\em Psychonomic Bulletin \& Review}, 21:785--793, 2014.

\bibitem{dubey2024llama}
Abhimanyu Dubey, Abhinav Jauhri, Abhinav Pandey, Abhishek Kadian, Ahmad Al-Dahle, Aiesha Letman, Akhil Mathur, Alan Schelten, Amy Yang, Angela Fan, et~al.
\newblock The llama 3 herd of models.
\newblock {\em arXiv preprint arXiv:2407.21783}, 2024.

\bibitem{geman1984stochastic}
Stuart Geman and Donald Geman.
\newblock Stochastic relaxation, gibbs distributions, and the bayesian restoration of images.
\newblock {\em {IEEE} Trans. Pattern Anal. Mach. Intell.}, 6(6):721--741, 1984.

\bibitem{griffiths2008using}
Thomas~L Griffiths, Brian~R Christian, and Michael~L Kalish.
\newblock Using category structures to test iterated learning as a method for identifying inductive biases.
\newblock {\em Cognitive Science}, 32(1):68--107, 2008.

\bibitem{griffiths2007language}
Thomas~L Griffiths and Michael~L Kalish.
\newblock Language evolution by iterated learning with bayesian agents.
\newblock {\em Cognitive science}, 31(3):441--480, 2007.

\bibitem{kalish2007iterated}
Michael~L Kalish, Thomas~L Griffiths, and Stephan Lewandowsky.
\newblock Iterated learning: Intergenerational knowledge transmission reveals inductive biases.
\newblock {\em Psychonomic Bulletin \& Review}, 14(2):288--294, 2007.

\bibitem{lewandowsky2009wisdom}
Stephan Lewandowsky, Thomas~L Griffiths, and Michael~L Kalish.
\newblock The wisdom of individuals: Exploring people's knowledge about everyday events using iterated learning.
\newblock {\em Cognitive science}, 33(6):969--998, 2009.

\bibitem{gpt4omini}
OpenAI.
\newblock Gpt-4o mini: advancing cost-efficient intelligence, 2024.
\newblock Accessed: 13-Sep-2024.

\bibitem{gpt4o}
OpenAI.
\newblock Hello gpt-4o, 2024.
\newblock Accessed: 13-Sep-2024.

\bibitem{reali2009evolution}
Florencia Reali and Thomas~L Griffiths.
\newblock The evolution of frequency distributions: Relating regularization to inductive biases through iterated learning.
\newblock {\em Cognition}, 111(3):317--328, 2009.

\bibitem{sanborn2007markov}
Adam Sanborn and Thomas Griffiths.
\newblock Markov chain monte carlo with people.
\newblock {\em Advances in neural information processing systems}, 20, 2007.

\bibitem{team2024gemma}
Gemma Team, Morgane Riviere, Shreya Pathak, Pier~Giuseppe Sessa, Cassidy Hardin, Surya Bhupatiraju, L{\'e}onard Hussenot, Thomas Mesnard, Bobak Shahriari, Alexandre Ram{\'e}, et~al.
\newblock Gemma 2: Improving open language models at a practical size.
\newblock {\em arXiv preprint arXiv:2408.00118}, 2024.

\bibitem{yamakoshi-etal-2022-probing}
Takateru Yamakoshi, Thomas Griffiths, and Robert Hawkins.
\newblock Probing {BERT}{'}s priors with serial reproduction chains.
\newblock In Smaranda Muresan, Preslav Nakov, and Aline Villavicencio, editors, {\em Findings of the Association for Computational Linguistics: ACL 2022}, pages 3977--3992, Dublin, Ireland, May 2022. Association for Computational Linguistics.

\bibitem{yeung2015identifying}
Saiwing Yeung and Thomas~L Griffiths.
\newblock Identifying expectations about the strength of causal relationships.
\newblock {\em Cognitive psychology}, 76:1--29, 2015.

\bibitem{zhu2024eliciting}
Jian-Qiao Zhu and Thomas~L Griffiths.
\newblock Eliciting the priors of large language models using iterated in-context learning.
\newblock {\em arXiv preprint arXiv:2406.01860}, 2024.

\bibitem{zhu2024recovering}
Jian-Qiao Zhu, Haijiang Yan, and Thomas~L Griffiths.
\newblock Recovering mental representations from large language models with markov chain monte carlo.
\newblock {\em arXiv preprint arXiv:2401.16657}, 2024.

\end{thebibliography}
\appendix
\section{Model Versions}
We use following snapshots for closed-source models: \texttt{gpt-4o-mini-2024-07-18}, \texttt{gpt-4o-2024-08-06}, \texttt{claude-3-haiku-20240307} and \texttt{claude-3-5-sonnet-20240620}.
\section{Prompts Used to Elicit LLM Priors}

We follow \citep{zhu2024eliciting} to prompt large language models. 
\paragraph{CoinFlip}
\begin{quote}
    \textbf{System}: Imagine that you are a participant in a psychology experiment. Your task is to evaluate the bias in a coin.
    
    \textbf{User}: Here is a brief overview of the past coin flips: Out of \texttt{Ntotal} coin flips, \texttt{Nhead} resulted in heads and \texttt{Ntotal}-\texttt{Nhead} in tails. With this information, please predict the number of heads in a larger set of 100 coin flips. Please limit your answer to a single value without outputting anything else.
\end{quote}

\paragraph{LifeExpectancy}
\begin{quote}
    \textbf{System}: You are an expert at predicting future events.

    \textbf{User}: If you were to evaluate the lifespan of a random \texttt{T}-year-old man, what age would you predict he might reach? Please limit your answer to a single value without outputting anything else.
\end{quote}

\section{Additional Results}

\begin{figure}[ht]
    \centering
    \begin{minipage}{0.32\textwidth}
        \centering
        \includegraphics[width=\linewidth]{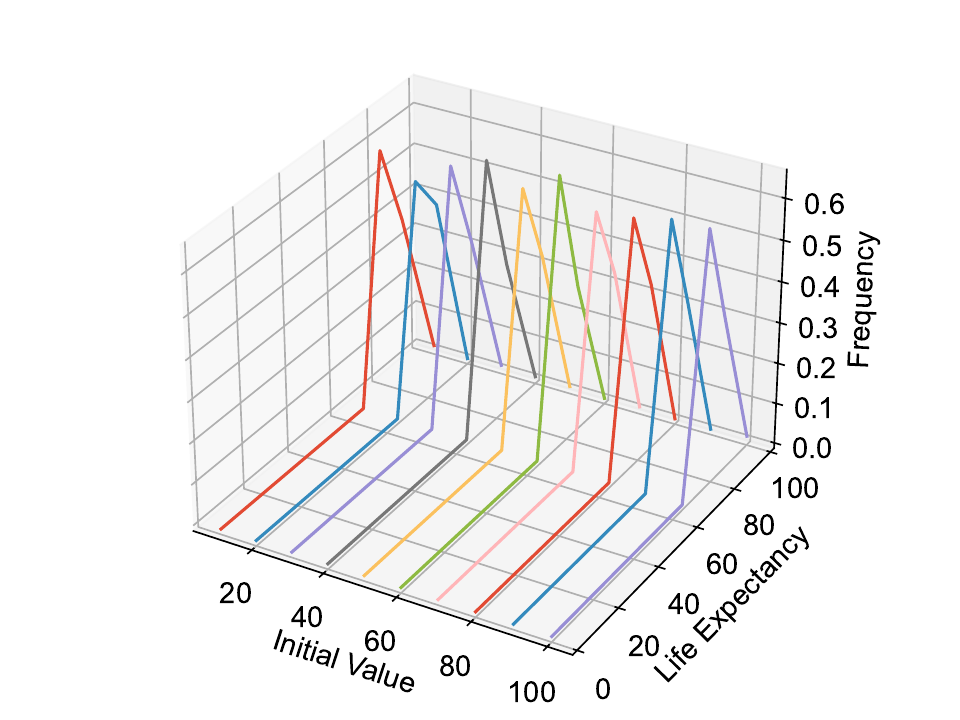}
        \subcaption{\texttt{claude-3-5-sonnet}}
    \end{minipage}
    \hfill
    \begin{minipage}{0.32\textwidth}
        \centering
        \includegraphics[width=\linewidth]{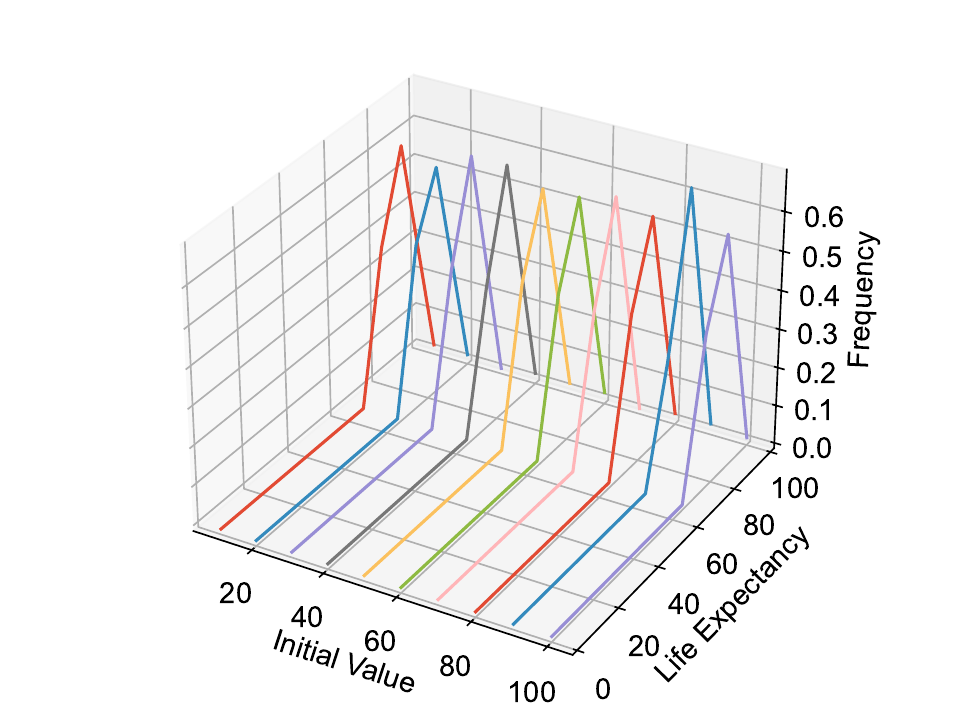}
        \subcaption{\texttt{gpt-4o}}
    \end{minipage}
    \hfill
    \begin{minipage}{0.32\textwidth}
        \centering
        \includegraphics[width=\linewidth]{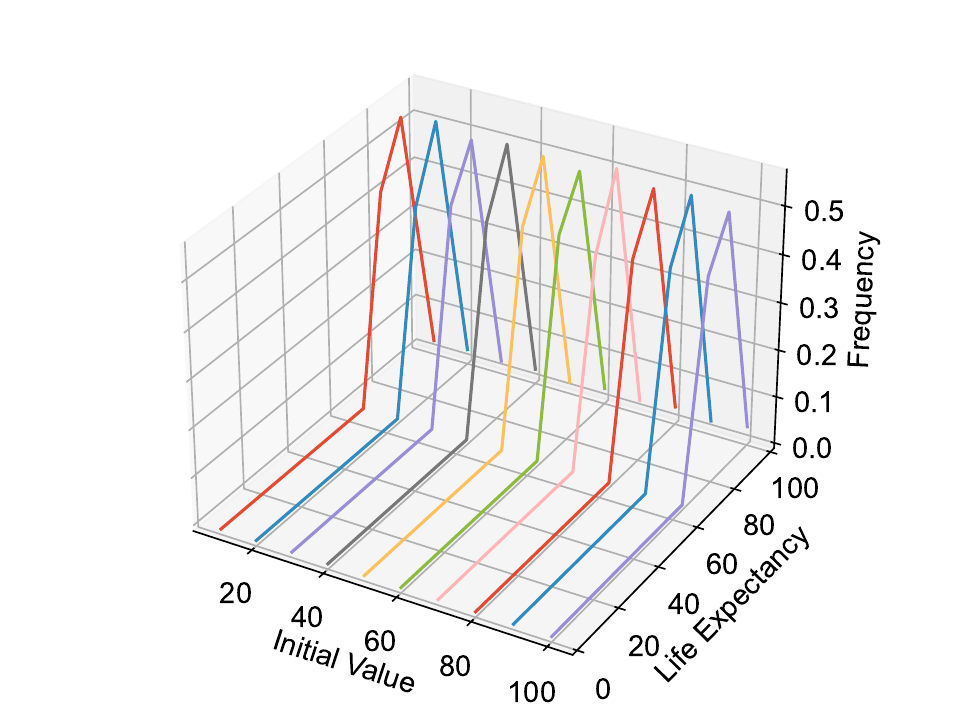}
        \subcaption{\texttt{Llama-3.1-70B-Instruct}}
    \end{minipage}
    \caption{Additional \textbf{LifeExpectancy} results.}\label{fig:life2}
\end{figure}

\end{document}